\documentclass[11pt]{article}

\usepackage[preprint]{acl}

\usepackage{xcolor}     
\usepackage{colortbl}  
\usepackage{times}
\usepackage{latexsym}
\usepackage[T1]{fontenc}
\usepackage[utf8]{inputenc}
\usepackage{microtype}
\usepackage{inconsolata}
\usepackage{graphicx}

\usepackage{booktabs}
\usepackage{multirow}
\usepackage{xcolor}
\usepackage{amsmath}
\usepackage{amssymb}
\usepackage{enumitem}
\usepackage{tabularx}
\usepackage{capt-of}  
\usepackage[capitalise,noabbrev]{cleveref}  
\usepackage{soul}

\hypersetup{
    colorlinks=true,
    linkcolor=red,   
}

\usepackage{pifont}          

\newcommand{\method}{\textbf{\texttt{ChunkAct}}}

\definecolor{gaincol}{RGB}{0, 140, 60}
\newcommand{\gain}[1]{{\scriptsize\bfseries\textcolor{gaincol}{#1}}}

\newcommand{\exitname}[1]{\texttt{#1}}
\newcolumntype{Y}{>{\raggedright\arraybackslash}X}  
\newcommand{\tool}[1]{\texttt{#1}}

\newcommand{\circnum}[1]{\raisebox{-1.1pt}{\ding{\numexpr181+#1\relax}}}

\title{\method: Action Chunking for Agentic Robotics}

\author{Bingxin Xu$^{1}$ \quad Wenrui Bao$^{2}$ \quad Xinxin Liu$^{2}$ \quad Jiawei Zhou$^{3}$ \quad Yuzhang Shang$^{2}$ \quad Emilio Ferrara$^{1}$ \\
$^{1}$University of Southern California \quad $^{2}$University of Central Florida \quad $^{3}$Stony Brook University}

\begin{document}
\maketitle

\begin{abstract}
Agentic robotics places a large language model (LLM) agent in the robot control loop, where it composes perception, motion, and vision-language-action (VLA) tools under the guidance of a memory and thereby generalizes to new tasks without retraining.
This generality comes at a steep price in efficiency.
Every primitive requires a round of LLM reasoning over the full interaction history, so the cost of an episode grows with the number of primitives.
We find this reasoning largely redundant: once a sub-goal is chosen, its primitives follow a routine already prescribed by memory.
We present \method{}, which introduces action chunking to agentic robotics.
Online, a single call launches a chunk of chained primitives, which executes autonomously and returns control only upon failure.
Offline, \method{} compiles the memory into these chunks, so execution is memory-free at test time and the planner is reserved for decisions that require reasoning.
LLM computation therefore scales with the number of decisions rather than the number of primitives.
On LIBERO-Pro, \method{} improves task success by 5.0 points over the state of the art in agentic robotics while requiring 2.9$\times$ fewer LLM calls, 4.4$\times$ fewer input tokens, and 2.3$\times$ fewer output tokens.
\end{abstract}

\section{Introduction}
\label{sec:intro}

\begin{figure*}[t]
  \centering
  \includegraphics[width=0.98\textwidth]{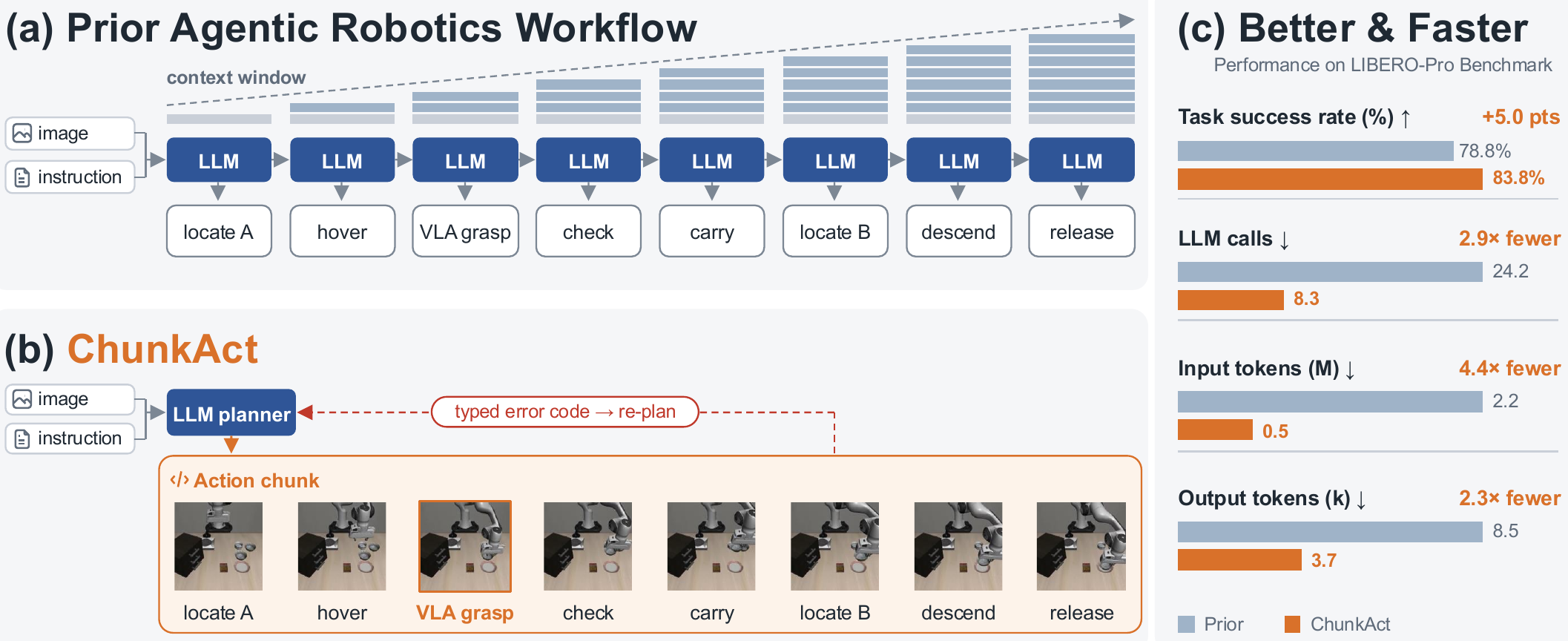}
  \vspace{-0.1in}
  \caption{\textbf{\method{} introduces action chunking to agentic robotics, improving both task success and efficiency.} Top: prior methods invoke the LLM planner once per primitive, so the number of LLM calls and the context length grow with every step. Bottom: \method{} invokes the planner once to launch an action chunk, which executes autonomously and returns a typed error only upon failure. Right: \method{} gains both task success and efficiency.}
  \label{fig:teaser}
  \vspace{-0.2in}
\end{figure*}

Agentic robotics has emerged as a new paradigm for robot manipulation \citep{liang2023cap,kumar2026aor,fu2026capx,tsui2026faea,zhang2026harnessvla,jia2026agentpolicy,xu2026baton,xu2026safeharness}. Rather than training a policy that maps pixels and language directly to motor commands \citep{brohan2023rt2,kim2024openvla,black2025pi0}, it places a large language model (LLM) agent in the control loop and equips it with perception, motion, and vision-language-action (VLA) tools, together with a memory of what has worked before. At each step, the agent observes the scene, consults the memory, reasons, and issues a single tool call (\cref{fig:teaser}, top). Because each task is composed by reasoning rather than learned from demonstrations, the agent handles new tasks without retraining and succeeds where end-to-end policies break down. For example, with no robot-specific training, GPT-6 Astra driven through a generic agent harness completes a real-arm pick-and-place task in 19 of 20 trials \citep{openai2026astra,robocurve2026astra} and ranks above every public policy on the RoboDojo benchmark \citep{zhang2026astraeval}, which illustrates the generalization this paradigm affords.

However, this generalization comes at a steep price in efficiency. The planner acts through primitives, i.e., robot skills invoked as tool calls, each of which requires a dedicated round of LLM reasoning. Every such round reprocesses the full interaction history, and reasoning models further consume output tokens before responding \citep{sui2025overthinking}. Consequently, the number of calls and output tokens scales linearly with the number of primitives in an episode, while input tokens scale quadratically \citep{kapoor2024agents,kim2026agenticcache}, as illustrated in Fig.\ref{fig:teaser} Top. For instance, a single pick-and-place task requires 24.2 LLM calls and 2.2M input tokens on average, costing about \$22 per episode with GPT-6 Astra and still \$11 even with the cheaper GPT-5.5.\footnote{Estimated at OpenAI's list API prices as of October 2026.} As tasks grow longer, this overhead quickly dominates the cost of execution, hindering the scalability of LLM-based agents.

To address this inefficiency, we analyze the agent's behavior and find that LLM reasoning is frequently redundant. Once a sub-goal is selected and its targets are localized, the subsequent primitives follow a fixed routine already encoded in memory, such that each additional call merely reproduces a decision that has already been made. Although a single read of memory suffices to capture this routine, the LLM tends to deliberate over each step and constantly returns to memory for guidance. Furthermore, the memory retrieval and LLM reasoning become less reliable as the context grows \citep{liu2024lostmiddle,laban2025multiturn}. This overhead could therefore be eliminated by compiling memory into generalizable action sequences that execute without the planner's step-by-step involvement. This concept also echoes the action chunk~\cite{zhao2023act} in robotics. 

Building on this observation, we propose \method{} (Action Chunking for Agentic Robotics), which consists of two complementary components: chunked execution at test time and memory internalization offline.
\textbf{At test time, \method{} introduces action chunking to agentic robotics.}
It composes the primitives that realize a sub-goal into an executable chunk, which the planner invokes with a single call and which then runs to completion autonomously.
This design confines the LLM to where reasoning is genuinely required: understanding the physical scene, diagnosing failures, and redirecting the plan.
To support this, a failed chunk returns a typed exit that specifies the failed primitive along with an error log, providing the planner with structured feedback for recovery.
In this way, LLM computation scales with the number of decisions rather than the number of primitives (See Fig.\ref{fig:teaser} Bottom).
Offline, \method{} builds these chunks as generalizable routines with memory compiled into them.
The parameters and procedural knowledge that govern each chunk are compiled from memory before deployment, making its actions inherently memory-conditioned.
As a result, chunks run memory-free at test time, eliminating frequent memory retrieval and further improving the efficiency of the LLM.

Together, these two components improve both efficiency and performance.
By substantially reducing the number of LLM calls and the length of the context, \method{} lowers the overall cost of each episode.
Meanwhile, because the planner reasons over a concise context and memory is internalized throughout execution, \method{} also achieves higher task success rates (See Fig.\ref{fig:teaser} Right). To summarize, our contributions are threefold.
\begin{itemize}[leftmargin=*,itemsep=1pt,topsep=2pt]
  \item To our knowledge, \method{} is the first LLM agent for robot manipulation that commits to a chunk of chained perception and control primitives per call rather than to a single primitive.
  \item \method{} compiles memory offline into generalizable action chunks, enabling memory-free execution at test time and reserving the LLM for decisions that genuinely require reasoning.
  \item Compared with the previous SOTA agentic robotics approach, \method{} improves success by 5.0 points on LIBERO-Pro, while requiring 2.9$\times$ fewer LLM calls, 4.4$\times$ fewer input tokens, and 2.3$\times$ fewer output tokens. It also exhibits strong zero-shot transfer capability.
\end{itemize}

\section{Related Work}
\label{sec:related}

\paragraph{Agentic robotics.}
Language models entered manipulation as planners that sequence a fixed repertoire of skills, grounded in affordances and closed by textual feedback \citep{ahn2022saycan,huang2022innermonologue}.
Code as Policies turned the plan into a program over perception and control interfaces \citep{liang2023cap}, and later work widened what the program could express with structured prompts, foundation perception, and value maps \citep{singh2023progprompt,huang2023instruct2act,huang2023voxposer,mu2024robocodex}.
A parallel line places a reasoning model above a learned policy rather than around it \citep{belkhale2024rth,shi2025hirobot,black2025pi05}.
The two lines have converged in \emph{agentic robotics}, where an LLM agent sits in the control loop, calls perception, motion, and VLA tools, and keeps a memory of what has worked \citep{fu2026capx,tsui2026faea,kumar2026aor,jia2026agentpolicy,zhang2026harnessvla,xu2026baton,xu2026safeharness}.
All of this work is measured by what the agent can accomplish, and none of it addresses the cost of consulting the planner before every primitive, which has been approached only by caching plans for tasks seen before \citep{zhang2025plancache,kim2026agenticcache}.
To our knowledge, we are the first to improve efficiency in agentic robotics systems by consolidating recurring call sequences into reusable action routines.

\paragraph{Action chunking in robotics.}
How many actions a policy commits to per decision is a design choice that trades reactivity against consistency and cost.
Policies that query the model once per control step \citep{brohan2022rt1,brohan2023rt2,kim2024openvla} tie inference cost to control frequency and let errors compound between queries.
Action chunking, introduced by ACT \citep{zhao2023act} and adopted by Diffusion Policy \citep{chi2023diffusionpolicy}, predicts a short sequence of actions per inference, and VLA models have since built it into their action heads and tokenizers \citep{black2025pi0,kim2025oft,pertsch2025fast}.
Subsequent work tunes the trade-off by interleaving chunks for closed-loop reactivity \citep{liu2024bid,black2025rtc}, and world-action models extend chunked prediction from actions to future observations \citep{du2023unipi,wu2023gr1,kim2026cosmospolicy}.
In all of these, a chunk is a short sequence of motor commands from a learned policy, and how much an LLM agent should commit to per decision has been asked only in text environments \citep{yang2026space}.
To our knowledge, we are \textit{the first to explore action chunking in agentic robotics, where a chunk spans perception, control primitives, and internalized memory rather than motor commands}.

\paragraph{Memory in agentic robotics.}
Memory lets an LLM agent improve without weight updates, and its forms differ in what is stored rather than in how it is used.
Reflective memory stores trajectories or the lessons drawn from them and returns relevant entries to the prompt \citep{park2023generative,shinn2023reflexion,zhao2024expel,packer2023memgpt}.
Procedural memory stores what worked as code, from skill libraries \citep{wang2023voyager} to induced workflows \citep{wang2024awm} and robot programs adapted to new contexts \citep{kagaya2025mtp,woods2026skillcomposer}.
Robot harnesses keep rule cards and task records written by the planner itself \citep{zhang2026harnessvla}, transition conditions between subtasks \citep{xu2026baton}, or a demonstration video navigated at a chosen granularity \citep{anonymous2026icl}.
In every form, the planner decides what to retrieve and when, reads it into context, and interprets it before acting, so the memory is consumed at test time and its value is bounded by the planner's ability to use it \citep{zhang2024memsurvey}.
\citet{li2025cdlm} comes closest to breaking this pattern by distilling recurring spans into a datastore that is emitted rather than read, yet its chunks remain text.
We are the first work to explore compiling agentic memory offline into physical action, producing executable action chunks that require no memory access at deployment.
\section{Action Chunking for Agentic Robotics}
\label{sec:method}

\subsection{Preliminaries}
\label{sec:preliminary}

We begin by formalizing agentic robotics, the setting in which \method{} operates.
\paragraph{The agentic loop.}
An agentic system solves a manipulation task with an LLM planner $\mathcal{G}$ that sequentially composes primitives from a library $\mathcal{P}$,
\begin{equation}
  \begin{gathered}
    a_t = (p_t, \theta_t) = \mathcal{G}(c_t), \qquad y_t = p_t(\theta_t), \\
    c_{t+1} = c_t \oplus (a_t, y_t).
  \end{gathered}
  \label{eq:loop}
\end{equation}
Here $c_t$ denotes the context window of the planner at step $t$, initialized with the task instruction as $c_0 = \ell$.
At each step, the planner reasons over $c_t$ and outputs $a_t$, which specifies the primitive to execute, $p_t \in \mathcal{P}$, and its arguments $\theta_t$.
The primitive runs to completion and returns a result $y_t$, which is appended to the context with $a_t$.
The process repeats until the task is completed at $T$.

Notably, not every planner output is a robot action.
Action primitives alter the physical state of the robot, whereas informational primitives leave the scene intact and only return content, either from perception or from a memory $\mathcal{M}$ (\cref{app:primitives}).
Since every result is appended to $c_t$, an informational call is paid for not only once, but again as input at every subsequent step.

As the planner re-reads its entire context at every call, a $T$-step episode costs $T$ planner calls and
\begin{equation}
  N_{\text{in}} = \sum_{t=0}^{T-1} |c_t|
  = T\,|c_0| + \sum_{s=0}^{T-2} (T-1-s)\,\bigl|(a_s, y_s)\bigr|
  \label{eq:cost}
\end{equation}
input tokens, where $|\cdot|$ counts tokens.
Each call reasons before it answers, so calls and output tokens grow linearly in $T$, while input tokens grow quadratically.
\Cref{eq:cost} identifies two quantities that govern the cost: the number of steps $T$, and the length of each appended pair $(a_s, y_s)$, which is paid again at every later step.
For instance, Harness VLA takes on average 24.2 planner calls and 2.2M input tokens per episode (see \cref{tab:efficiency}).

\begin{figure}[t]
  \centering
  \vspace{-0.15in}
  \includegraphics[width=0.77\columnwidth]{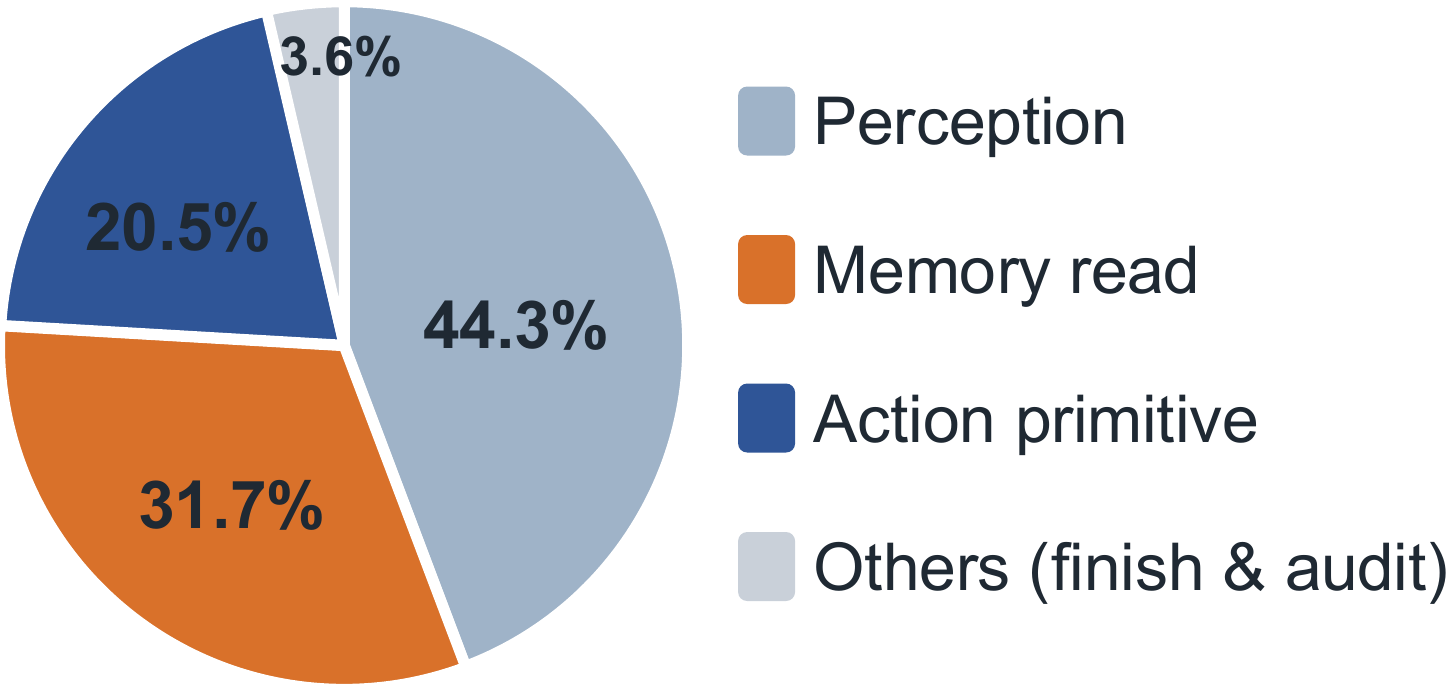}
  \vspace{-0.1in}
  \caption{\textbf{Distribution of LLM planner's tool calls on LIBERO.} Memory reads and perception dominate.}
  \vspace{-0.2in}
  \label{fig:pie}
\end{figure}

We analyze the planner's tool calls and reveal two sources of inefficiency.
\circnum{1} \ul{The planner conflates deciding with executing.}
Every primitive incurs a full round of reasoning, even though, once a sub-goal is fixed, the primitives that realize it are largely determined by the memory; the planner is thus invoked far more often than the task demands decisions.
\circnum{2} \ul{The planner spends most of its calls gathering information rather than acting.}
Only 20.5\% of calls invoke an action primitive, while memory reads and perception account for 31.7\% and 44.3\%, respectively (See Fig.\ref{fig:pie}).
Since $\mathcal{M}$ is static across episodes, these reads retrieve the same knowledge again and again, inflating the context without informing new decisions.
These two findings motivate \method{}, which \textbf{brings action chunking to agentic robotics}  through offline action chunk building (\cref{sec:distill}) and online action chunk triggering (\cref{sec:protocol}).

\begin{figure*}[t]
  \centering
  \includegraphics[width=\textwidth]{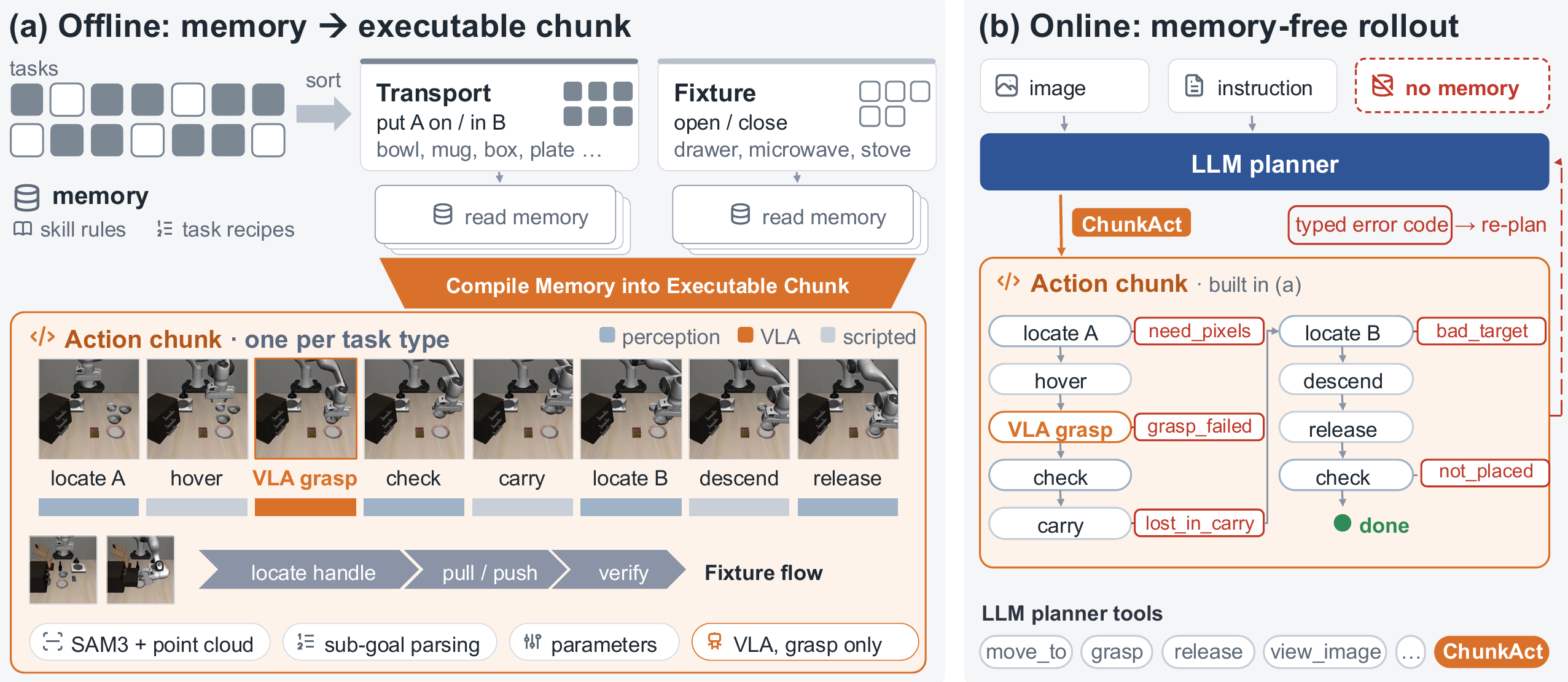}
  \vspace{-0.2in}
  \caption{\textbf{The two stages of \method{}.} Left: offline, \method{} internalizes memory by compiling it into executable action chunks. Right: online, the LLM planner invokes \method{} via a single function call, which executes an entire action chunk autonomously and returns a typed error code to the planner only upon failure.}
  \vspace{-0.2in}
  \label{fig:pipeline}
\end{figure*}

\subsection{Offline action chunk building}
\label{sec:distill}

Targeting the \circnum{1} issue, \method{} outputs action chunks rather than single primitives.
The planner is reengaged only when the chunk fails, reserving LLM reasoning for the decisions that require it.

The memory of an LLM agent already records the general, reusable success rules and and procedural context for tasks. \method{} compiles all memory once into executable chunk form rather than letting the planner re-derive it at every step (\cref{fig:pipeline}(a)).
Memory construction begins with a partition over task categories. A manipulation instruction decomposes into an ordered sequence of typed sub-goals, and the space of such types remains small even as the task set grows. For example, two types suffice to cover a broad range of tabletop manipulation: transport, which relocates an object onto or into a receptacle, and fixture, which actuates an articulated element of the scene. All memory entries pertaining to a given type, from general execution rules to task-specific records, are therefore consolidated into a single source and compiled into a unified chunk, parameterized by the object classes on which it operates.

Compilation retains only the procedural knowledge that defines how a task type is executed: the ordering of stages, the primitive assigned to each stage, the stage-level parameters conditioned on object class, the fallback strategies to attempt when a stage fails, and the criteria used to verify each stage. The primitive assignment, for instance, preserves the division of labor recorded in memory, delegating contact-rich stages such as grasping or fixture actuation to the VLA policy and the remaining stages to analytic primitives. Conversely, any quantity unobservable at deployment time is excluded. Recorded coordinates and privileged simulator state are discarded, and all geometric quantities are re-estimated from live perception of the current scene. This decoupling allows a chunk to transfer to scenes whose layouts differ from those encountered during memory construction.

The resulting chunk is an executable routine that captures the recurring procedure of its task type. A single chunk covers every task of its type, with class-dependent content resolved by lookup within the chunk. The planner is thus relieved of both memory retrieval and step-by-step primitive selection at run time. During execution, each stage is verified before the next begins, and the library is compiled once and kept frozen.

\subsection{Online action chunk triggering}
\label{sec:protocol}

To sovle the \circnum{2} problem, \method{} internalizes the memory $\mathcal{M}$ into executable chunks, allowing memory-free execution at run time.

At test time the planner invokes chunks in place of primitives, through a single tool call (\cref{fig:pipeline}(b) labels \method{}).
The instruction is parsed into its ordered sequence of typed sub-goals, each sub-goal selects the chunk of its kind, and the planner's interface to execution narrows to two operations, grounding a target in the current observation and launching the chunks.
An episode therefore begins with the planner interpreting the task and the scene, grounding the first target, and issuing one call, after which the chunks run the sub-goals until the task is complete without any further request to the planner.
No memory is read along the way, since everything the planner would have retrieved has been compiled into the chunks, and the planner's context holds only the instruction, the observations it has requested, and the exits it has received.

Chaining the stages under one call, rather than exposing each stage as a separate call, follows from where reasoning is actually needed.
Compiling memory and perception into chunks removes the cost of retrieval and repeated observation, but a per-stage interface would still return control to the planner after every action, turning each step of execution into a round of deliberation.
Most of these rounds involve no decision.

Once its target is grounded, the stages of a sub-goal follow a fixed order, so a successful stage already determines the next one and requires no reasoning to select it. A chunk therefore chains the stages directly and returns control to the planner only when a stage fails.

Control returns to the planner only when a chunk fails, and it does so through a typed exit.
Every stage of a chunk raises its own error code, so the exit names the stage that failed and the reason it failed, such as a perception call that cannot locate its target, a policy call that loses the object, or a verification that cannot confirm the sub-goal, and it carries the error log with the robot state and fresh observations (\cref{app:exits}).
The planner then re-perceives the scene from its true state, and chooses among three responses, namely grounding the target again and re-invoking the chunk, redoing a sub-goal whose verification failed, or falling back to the primitives for a situation the library does not cover.
Each response is a decision that requires understanding the task and the scene, and none is a step of execution, so the planner is consulted at decision points and nowhere else.

\begin{table*}[t]
\centering
\footnotesize
\setlength{\tabcolsep}{5pt}
\caption{\textbf{Task success on LIBERO-Pro.} Task success rate (\%) under instruction redirection (T) and position swap (S) for each suite. Harness VLA and \method{} both use GPT-5.6-sol with high thinking. Best entry is in bold.}
\vspace{-0.1in}
\label{tab:main}
\scalebox{0.95}{
\begin{tabular}{@{}cl ccccccccc@{}}
\toprule
& \textbf{Method} & \textbf{Spat-T} & \textbf{Spat-S} & \textbf{Obj-T} & \textbf{Obj-S} & \textbf{Goal-T} & \textbf{Goal-S} & \textbf{L10-T} & \textbf{L10-S} & \textbf{Overall} \\
\midrule
\multirow{8}{*}{\rotatebox[origin=c]{90}{VLA}}
& OpenVLA \citep{kim2024openvla}        & 0.0  & 0.0  & 0.0  & 0.0  & 0.0  & 0.0  & 0.0  & 0.0  & 0.0 \\
& $\pi_0$ \citep{black2025pi0}          & 0.0  & 0.0  & 0.0  & 2.0  & 0.0  & 0.0  & 0.0  & 0.0  & 0.3 \\
& $\pi_{0.5}$ \citep{black2025pi05}     & 1.0  & 20.0 & 1.0  & 17.0 & 2.0  & 38.0 & 1.0  & 8.0  & 11.0 \\
& MolmoAct \citep{lee2025molmoact}      & 0.0  & 0.0  & 0.0  & 6.0  & 0.0  & 0.0  & 6.0  & 0.0  & 1.5 \\
& NORA \citep{hung2025nora}             & 0.0  & 0.0  & 0.0  & 0.0  & 0.0  & 0.0  & 0.0  & 0.0  & 0.0 \\
& X-VLA \citep{zheng2025xvla}           & 0.0  & 0.0  & 8.0  & 2.0  & 9.0  & 1.0  & 10.0 & 0.0  & 3.8 \\
& AtomVLA \citep{sun2026atomvla}        & 1.0  & 16.0 & 0.0  & 10.0 & 11.0 & 2.0  & 9.0  & 1.0  & 6.3 \\
& $\pi_{\mathrm{RLinf}}$ \citep{yu2025rlinf} & 42.0 & 59.0 & 71.0 & 78.0 & 45.0 & 42.0 & 49.0 & 14.0 & 50.0 \\
\midrule
& CaP-X \citep{fu2026capx}               & 14.0 & 12.0 & 18.0 & 22.0 & 17.0 & 26.0 & /    & /    & 18.2 \\
& Harness VLA \citep{zhang2026harnessvla} & \textbf{100.0} & 80.0 & 80.0 & \textbf{100.0} & 70.0 & 80.0 & \textbf{60.0} & 60.0 & 78.8 \\
\rowcolor{blue!15}
\multirow{-3}{*}{\rotatebox[origin=c]{90}{Agent}}
& + \method{} (ours) & 94.0 & \textbf{82.0} & \textbf{98.0} & \textbf{100.0} & \textbf{80.0} & \textbf{86.0} & \textbf{60.0} & \textbf{70.0} & \textbf{83.8} \\
\bottomrule
\end{tabular}}
\vspace{-0.2in}
\end{table*}

\subsection{Role of LLM planner in agentic robotics}
\label{sec:analysis}

We examine how this design changes the role of the planner, from stepwise executor to sub-goal selector and failure handler.

Taken together, the two stages redefine the role of the planner.
In the per-primitive loop of \cref{eq:loop}, the planner serves as both reasoner and executor, treating every step alike regardless of whether it requires judgment.
\method{} separates the two: execution is delegated to the chunks, while the planner retains only what cannot be compiled, namely understanding the task and scene and responding to failure.
This separation reshapes both sides of the loop.

On the input side, the dominant sources of context are removed.
Memory is internalized in the chunks rather than retrieved online, and perception is performed within each chunk rather than requested before every primitive.
The context thus stays compact across an episode, containing only the instruction, decision-time observations, and exits.
On the output side, the unit of action shifts from a single primitive to a chained chunk.
Each step $a_t$ is either a chunk launch or a response to an exit, so the planner is invoked per decision rather than per action, and reasoning is spent only where it is required.
Episode cost therefore scales with the number of decisions rather than procedure length.

\section{Experiments}
\label{sec:exp}

\subsection{Setup}
\label{sec:setup}

\begin{table}[t]
\centering
\small
\setlength{\tabcolsep}{3.5pt}
\caption{\textbf{Efficiency analysis and LLM role ablation on LIBERO-Pro.} Average TSR (\%), planner calls, input tokens (M), and output tokens (k) with GPT-5.6-sol in two reasoning settings. Green rows show the gain over Harness VLA. Last row runs chunks with no LLM call.}
\label{tab:efficiency}
\begin{tabular}{@{}llcccc@{}}
\toprule
\textbf{Method} & \textbf{Reasoning} & \textbf{TSR} & \textbf{Calls} & \textbf{In (M)} & \textbf{Out (k)} \\
\midrule
Harness VLA & High & 78.8 & 24.2 & 2.2 & 8.5 \\
+\method{}   & High & \textbf{83.8} & \textbf{8.3} & \textbf{0.5} & \textbf{3.7} \\
 & & \gain{+5.0} & \gain{2.9$\times$} & \gain{4.4$\times$} & \gain{2.3$\times$} \\
\midrule
Harness VLA & None & 56.0 & 24.9 & 2.4 & 6.4 \\
+\method{}   & None & \textbf{80.0} & \textbf{6.6} & \textbf{0.4} & \textbf{1.3} \\
 & & \gain{+24.0} & \gain{3.8$\times$} & \gain{6.0$\times$} & \gain{4.9$\times$} \\
\midrule
+\method{}   & No planner & 57.5 & / & / & / \\
\bottomrule
\end{tabular}
\end{table}

\noindent\textbf{Benchmarks.}
We evaluate on two benchmarks from the LIBERO family.
Our main experiments use LIBERO-Pro \citep{zhou2025liberopro}, which perturbs the four LIBERO suites \citep{liu2023libero}, Spatial, Object, Goal, and LIBERO-10, along objects, initial states, instructions, and environments. Following \citet{zhang2026harnessvla}, we evaluate the instruction and initial-state perturbations, denoted instruction redirection (T) and position swap (S).
Each of the resulting eight suite and perturbation combinations contains 10 tasks.
As in that protocol, we run 10 seeds per task, so that each combination aggregates 800 episodes.
We further use LIBERO-Plus \citep{fei2025liberoplus} benchmark, which varies the LIBERO tasks along seven axes: object layout, camera viewpoint, robot initial state, language, lighting, background, and sensor noise. 
As its 10,030 variants are too many to evaluate exhaustively, we run a single seed on every task under each of the seven perturbation axes, which covers the full range of perturbations at a tractable cost.

\paragraph{Metrics.}
We measure performance by the task success rate (TSR), as determined by the environment's termination signal.
We measure efficiency by the number of planner calls and the numbers of input and output tokens per episode. 

\paragraph{Baselines.}
For the main experiment, we compare with eight end-to-end VLA models, OpenVLA \citep{kim2024openvla}, $\pi_0$ \citep{black2025pi0}, $\pi_{0.5}$ \citep{black2025pi05}, MolmoAct \citep{lee2025molmoact}, NORA \citep{hung2025nora}, X-VLA \citep{zheng2025xvla}, AtomVLA \citep{sun2026atomvla}, and the RLinf-fine-tuned $\pi_{0.5}$ checkpoint that serves as our VLA primitive, which we denote $\pi_{\mathrm{RLinf}}$ \citep{yu2025rlinf}, and with CaP-X \citep{fu2026capx}, an agent that writes code over perception and motion primitives without a VLA.
Their results are those reported by \citet{zhang2026harnessvla}.
Harness VLA and \method{} are run by us. For LIBERO-Plus experiment results, the VLA models' performance are as reported by \citet{fei2025liberoplus}.

\begin{table*}[t]
\centering
\small
\setlength{\tabcolsep}{4pt}
\caption{\textbf{Task success on LIBERO-Plus.} Task success rate (\%) under all seven perturbation setting.  \method{} runs GPT-5.6-sol at the high reasoning setting with seed 0. The best entry in each column is in bold.}
\label{tab:plus}
\vspace{-0.1in}
\begin{tabular}{@{}lcccccccc@{}}
\toprule
\textbf{Method} & \textbf{Camera} & \textbf{Robot} & \textbf{Language} & \textbf{Light} & \textbf{Background} & \textbf{Noise} & \textbf{Layout} & \textbf{Overall} \\
\midrule
OpenVLA \citep{kim2024openvla}            & 1.1  & 4.1  & 26.8 & 4.4  & 25.3 & 19.3 & 31.6 & 16.1 \\
OpenVLA-OFT \citep{kim2025oft}            & 59.7 & 37.2 & 81.5 & 85.8 & 92.4 & 76.7 & 77.1 & 72.9 \\
OpenVLA-OFT$_{w}$ \citep{kim2025oft}      & 16.8 & 43.7 & 73.2 & 68.2 & \textbf{92.5} & 51.4 & 72.3 & 59.7 \\
OpenVLA-OFT$_{m}$ \citep{kim2025oft}      & 57.9 & 30.6 & 83.6 & 91.6 & 83.6 & 76.3 & 73.2 & 71.0 \\
$\pi_0$ \citep{black2025pi0}              & 15.8 & 6.6  & 61.0 & 79.6 & 78.5 & 79.4 & 70.4 & 55.9 \\
$\pi_0$-FAST \citep{pertsch2025fast}      & 66.4 & 24.8 & 63.3 & 73.0 & 67.7 & 75.8 & 70.3 & 63.0 \\
NORA \citep{hung2025nora}                 & 4.0  & 41.1 & 67.0 & 31.0 & 50.5 & 17.6 & 63.9 & 39.3 \\
WorldVLA \citep{cen2025worldvla}          & 0.3  & 30.2 & 44.2 & 29.4 & 14.5 & 12.2 & 39.4 & 24.3 \\
UniVLA \citep{bu2025univla}               & 4.3  & 50.3 & 71.8 & 59.1 & 80.0 & 25.3 & 34.3 & 46.4 \\
RIPT-VLA \citep{tan2025ript}              & 58.3 & 36.7 & 80.1 & 87.9 & 90.4 & 73.8 & 76.5 & 72.0 \\
\midrule
\rowcolor{blue!15}+ \method{} (ours)                          & \textbf{77.5} & \textbf{75.0} & \textbf{87.5} & \textbf{95.0} & 82.5 & \textbf{87.5} & \textbf{82.5} & \textbf{83.9} \\
\bottomrule
\end{tabular}
\end{table*}

\paragraph{Implementation.}
The planner is GPT-5.6-sol at high reasoning effort running in Codex \citep{openai2025codex}, with a budget of 100 LLM calls per episode.
The VLA primitive is $\pi_{\mathrm{RLinf}}$, and SAM~3 \citep{carion2025sam3} provides segmentation.
\method{} and Harness VLA \citep{zhang2026harnessvla} share the same set-up and the same task-specific and global memory.
They differ only in how the memory is used: Harness VLA reads it at test time, whereas \method{} internalizes it offline into executable chunks and reads no explicit memory at test time.

\subsection{Main results}
\label{sec:main}

\noindent\textbf{Task success performance.}
As shown in \cref{tab:main}, \method{} achieves the highest task success rate on LIBERO-Pro, outperforming both end-to-end VLAs and agentic robotic systems.
It attains an overall success rate of 83.8\%, exceeding the strongest prior method, Harness VLA, by 5.0 points.
The perturbations in LIBERO-Pro, instruction redirection and position swap, require a policy to act on instructions and object layouts that differ from those seen in training, and end-to-end VLAs largely fail under them.
This failure extends to $\pi_{\mathrm{RLinf}}$, the checkpoint that \method{} invokes as its contact primitive, which trails \method{} by 43.8 points.
Since the primitive alone does not generalize to these perturbations, whatever success an agent recovers must come from how it grounds the task in the current scene before the primitive is called.
Placing an LLM planner above the VLA recovers much loss, as the agentic baselines show, yet Harness VLA stops at 78.8\% despite sharing the same planner, primitive, perception, and memory with \method{}.
The two systems differ only in how memory is used.
Harness VLA retrieves memory at test time and relies on the planner to reinterpret past experience at every call, whereas \method{} compiles it offline into executable chunks that encode how each task succeeds.
The gap therefore isolates the benefit of memory internalization.

\noindent\textbf{Efficiency analysis.}
As shown in \cref{tab:efficiency}, \method{} attains its higher success rate at a fraction of the LLM cost.
The first two rows correspond to the Harness VLA and \method{} entries in \cref{tab:main}.
Relative to Harness VLA, \method{} issues 2.9$\times$ fewer LLM calls and consumes 4.4$\times$ fewer input tokens and 2.3$\times$ fewer output tokens, while improving success by 5.0 points.
The third and fourth rows repeat this comparison with reasoning disabled, and the efficiency advantage of \method{} grows further.
Both observations stem from memory internalization.
Harness VLA spends most LLM calls on perception queries and memory reads, and each result is appended to the context, so the context keeps growing over the episode.
\method{} folds these steps into the executable chunks, so they run at test time without invoking the LLM and their results never enter the context window.
The planner is called only at the decision points that the chunks leave open.
Fewer calls over a shorter context reduce the input tokens, and fewer planning decisions reduce the output tokens.

\begin{figure*}[t]
  \centering
  \includegraphics[width=\textwidth]{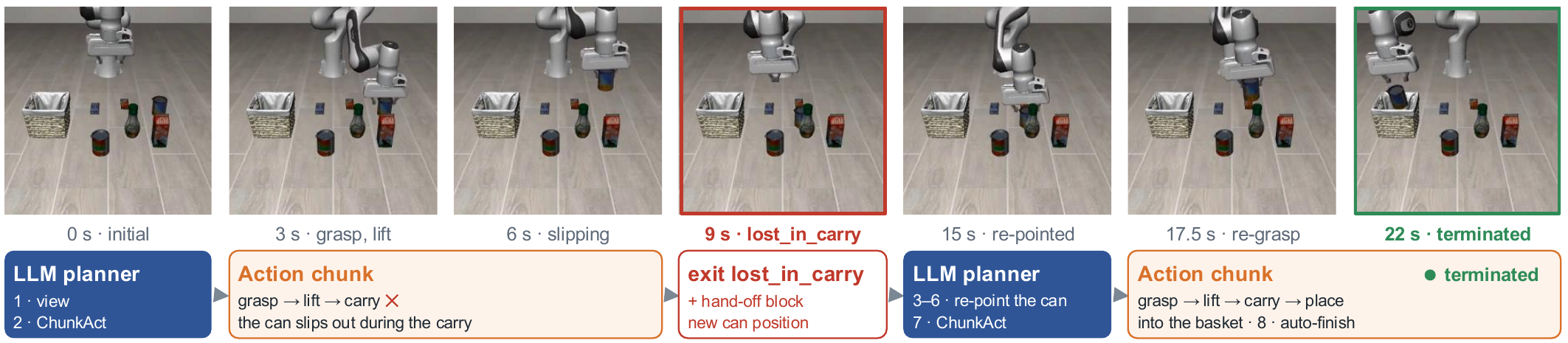}
\vspace{-0.2in}
  \caption{\textbf{Demo of a successful rollout on LIBERO-Pro with \method{}.} Given the task ``pick up the can and place it in the basket,'' the LLM planner first perceives the scene and invokes ChunkAct to execute a chained action sequence. The can is grasped successfully but slips during transport, so ChunkAct returns the error code. The planner interprets this feedback, re-localizes the can, and invokes ChunkAct again, which completes the task.}
\vspace{-0.2in}
  \label{fig:demo}
\end{figure*}

\subsection{Analysis of LLM planner role in \method{}}
\label{sec:planner}

\noindent\textbf{Analyzing the LLM planner reasoning impact.}
As shown in \cref{tab:efficiency}, \method{} remains robust when the reasoning of LLM planner is disabled, whereas Harness VLA does not.
The task success rate of \method{} stays stable across reasoning settings, while Harness VLA degrades sharply once reasoning is turned off.
This contrast reflects how heavily each system relies on the planner.
Harness VLA depends on the LLM at every step: each call must interpret the retrieved memory, decide what to perceive, and issue the correct next action, so the errors of a weaker planner accumulate over the episode.
\method{} instead decouples the LLM from the execution it does not need to perform.
With memory already internalized into the chunks, the planner is consulted only for the remaining decisions, which leaves a weaker planner far fewer opportunities to fail.

\noindent\textbf{Quantifying the LLM agent's contribution.}
To measure how much the planner contributes within \method{}, the last row of \cref{tab:efficiency} executes the chunks under a scripted driver that makes no LLM call.
The chunks alone reach 57.5\%, which corresponds to episodes in which the compiled routines succeed without intervention.
The LLM in \method{} without reasoning therefore accounts for the remaining 22.5 points.
This contribution comes from recovery: when a chunk fails, the planner is invoked to interpret the current scene and direct the next step, which a scripted driver cannot.
The LLM thus serves not as a step-by-step executor but as a recovery mechanism invoked where the compiled routine falls short, which is why it adds substantial success at a small cost in calls and tokens.

\subsection{Generalization to unseen perturbations}
\label{sec:plus}

\textbf{Out-of-distribution test.}
The memory of \method{} is compiled on LIBERO-Pro and includes task-specific memory, which prior work found necessary on that benchmark~\citep{zhang2026harnessvla}.
To test whether \method{} overfits to these tasks, we apply it unchanged to LIBERO-Plus~\citep{fei2025liberoplus}, which perturbs the LIBERO tasks along seven dimensions that the memory doesn't cover.
The chunk library and planner prompt are those of \cref{sec:main}, so the evaluation is fully zero-shot, whereas the VLA baselines, whose numbers are taken from \citet{fei2025liberoplus}, are fine-tuned on the benchmark data.

\noindent\textbf{Zero-shot transfer.}
\method{} achieves the best overall success rate in \cref{tab:plus}, exceeding the strongest prior method by 11.0 points and leading in six of the seven dimensions.
The margin is widest under camera and robot initial-state perturbations, where \method{} stays above three quarters while the best VLA falls to two thirds and one half, respectively.
Because a chunk recomputes its targets from perception in every episode, such changes alter the measured geometry but not the procedure.
These results show zero-shot transfer of internalized memory  to unseen perturbations rather than overfitting to the tasks it was compiled from.

\subsection{Qualitative analysis}
\label{sec:qualitative}

\cref{fig:demo} follows a single LIBERO-Pro episode and shows how the planner's eight calls are distributed over 22 seconds of execution.
The planner is engaged at two points only.
At the start it views the scene and launches a chunk, which grasps and lifts the can without further LLM request.
During the move, however, the can slips out of the gripper, and the chunk terminates with a typed exit.
It returns a hand-off block, a structured error report that adds context to the error code.
The planner resumes from the true state of the scene rather than from its expectation of it, spends four calls re-pointing the can, and triggers ChunkAct again, which carries the can to the basket, places it, and finishes on its own once the environment reports success.
The episode makes visible the division of labor that \cref{tab:efficiency} quantifies.
The planner is reserved for understanding, while the chunks handle execution.
It plans the next action chunk from the scene, and re-plans when a chunk fails.

\vspace{-0.05in}
\section{Conclusion}
\label{sec:conclusion}
\vspace{-0.05in}

Agentic robotics generalizes to novel tasks without retraining, but at a steep cost in efficiency. We identify two bottlenecks: the LLM planner reasons before every primitive, although the primitives of a sub-goal follow a recurring pattern, and most of its calls are spent on perception and memory retrieval rather than action. \method{} is the first work to introduce action chunking to agentic robotics, and it internalizes memory into executable chunks offline. The LLM is thereby decoupled from routine execution and engaged only where reasoning is genuinely required, which improves LLM efficiency significantly. On LIBERO-Pro, \method{} improves task success by 5.0 points over the SOTA method with 2.9$\times$ fewer LLM calls, 4.4$\times$ fewer input tokens, and 2.3$\times$ fewer output tokens, and it transfers zero-shot to unseen tasks on LIBERO-Plus.

\bibliography{references}

\appendix

\twocolumn[%
\begin{center}
\small
\captionof{table}{\textbf{The complete taxonomy of typed exits.} Each exit names the reason a chunk stopped and the reaction expected from the planner.}
\label{tab:exits}
\begin{tabularx}{\textwidth}{@{}llY@{}}
\toprule
\textbf{Group} & \textbf{Exit} & \textbf{Meaning and expected planner reaction} \\
\midrule
Success & \exitname{terminated} & The environment reported success inside the chunk; the finish is recorded and the planner session interrupted. \\
 & \exitname{placed\_subgoal} & A non-final sub-goal was placed and verified; the chunk continues with the next sub-goal (internal). \\
 & \exitname{done\_covered} & Every covered sub-goal was executed but the task did not terminate; re-perceive and continue. \\
\midrule
Hand-off & \exitname{need\_pixels} & The next object was not found by name, or the pixels back-projected off the workspace; call \tool{approach} with pixels on it. \\
 & \exitname{need\_receptacle\_pixels} & The target region cannot be derived (push or beside, or the receptacle point is implausibly high); pass \textit{receptacle\_pixels}. \\
 & \exitname{receptacle\_rejected} & The receptacle was not found by name, lies at an implausible height, or the fixture geometry was not located; pass \textit{receptacle\_pixels} or use primitives. \\
 & \exitname{wrong\_object\_under\_gripper} & The wrist camera sees the nearest object more than 5\,cm from the pointed spot; confirm or re-point. \\
 & \exitname{subgoal\_unverified} & The object was released but perception could not confirm the predicate; check the image, then redo or continue. \\
\midrule
Manipulation & \exitname{grasp\_failed} & Up to four grasp attempts, nothing held. \\
 & \exitname{lost\_at\_lift} & The object slipped during the lift, twice. \\
 & \exitname{lost\_in\_carry} & The object slipped during the carry, leveling, pitched reach, or hover re-centering. \\
 & \exitname{lost\_in\_descent} & The object left the gripper while lowering. \\
 & \exitname{carry\_stalled} & The arm stopped short of the point above the receptacle (reach limit). \\
 & \exitname{delivered\_unverified} & The VLA carried the object away and released it; the gripper is clear. \\
 & \exitname{placed\_not\_terminated} & The object was carried and released at the receptacle, but the task did not terminate. \\
\midrule
Fixture & \exitname{open\_unverified} & The drawer handle did not move after the contact skill. \\
 & \exitname{not\_terminated} & The close, turn-on, or door-sweep skill ran but the task did not terminate. \\
\midrule
Scope & \exitname{not\_covered} & The sub-goal or placement is not covered by any chunk; continue with primitives. \\
 & \exitname{error} & Internal error, for instance no pixel can be back-projects; continue with primitives. \\
\bottomrule
\end{tabularx}
\end{center}
\vspace{1em}
]

\section{Typed Exit Taxonomy}
\label{app:exits}
\cref{tab:exits} lists all exit codes that a chunk may return upon failure, with its meaning and expected planner reaction.

\section{Primitive Library}
\label{app:primitives}

\begin{table}[t]
\centering
\small
\setlength{\tabcolsep}{4pt}
\caption{\textbf{The primitive library $\mathcal{P}$.} Primitives of Harness VLA~\citep{zhang2026harnessvla}. The last row lists the single call that \method{} adds to this library.}
\label{tab:primitives}
\begin{tabularx}{\columnwidth}{@{}lY@{}}
\toprule
\textbf{Type} & \textbf{Primitives} \\
\midrule
Perception & \texttt{view\_env\_state}, \texttt{segment}, \texttt{back\_project}, \texttt{view\_camera\_meta}. Return the current images and robot state, segment by text or point prompt, and back-project pixels to 3D points. \\
\addlinespace
Motion & \texttt{move\_to}, \texttt{move\_pose}, \texttt{set\_gripper}, \texttt{release}, \texttt{rotate\_pitch}, \texttt{rotate\_wrist}. These primitives are executed in closed loop by an analytic controller until the target pose or gripper state is reached. \\
\addlinespace
VLA & \texttt{pi0\_pick}, \texttt{pi0\_doubled}. Drive the frozen $\pi_{0.5}$ policy under a language command until a lift or stop condition is met. \\
\addlinespace
Memory & \texttt{read\_text\_file}, \texttt{list\_dir}, \texttt{write\_text\_file}. Read memory cards and guides from $\mathcal{M}$ and write audit notes. \\
\addlinespace
Control & \texttt{finish}. Declares the task complete. \\
\midrule
\method{} & \texttt{approach}, which runs an action chunk. \\
\bottomrule
\end{tabularx}
\end{table}

\cref{tab:primitives} lists all the primitives that constitute the library $\mathcal{P}$ introduced in \cref{sec:preliminary}, which are employed in all experiments of \cref{sec:exp}.
\method{} retains this library without modification and adds one call, shown in the last row.

\end{document}